\documentclass[conference]{IEEEtran}
\usepackage{amsmath}
\usepackage{graphicx}

\usepackage{multirow}
\usepackage{booktabs}
\usepackage{xcolor,colortbl}
\usepackage{arydshln}

\definecolor{Gray}{gray}{0.85}
\definecolor{LightCyan}{rgb}{0.88,1,1}

\newcolumntype{a}{>{\columncolor{Gray}}c}
\newcolumntype{b}{>{\columncolor{white}}c}

\usepackage{xcolor,soul,framed} 
\usepackage{color}
\usepackage[normalem]{ulem}

\colorlet{shadecolor}{yellow}

\usepackage{eqparbox}
\usepackage{multirow}
\usepackage{url}

\usepackage[left=0.75in,right=0.75in,top=0.75in,bottom=0.75in]{geometry}

\begin{document}

\bstctlcite{IEEEexample:BSTcontrol}
  \title{\vspace*{0.25in} Assessing Cross-dataset Generalization of Pedestrian Crossing Predictors}
  %
\author{\parbox{17cm}{\centering
    {\large Joseph Gesnouin$^{1,2}$, Steve Pechberti$^{1}$, Bogdan Stanciulescu$^{2}$ and Fabien Moutarde$^{2}$}\\
    {\normalsize
    $^1$ Institut VEDECOM, 78000 Versailles, France\\
    $^2$ Centre de Robotique, MINES ParisTech, Université PSL, 75006 Paris, France}}
}



\maketitle

\begin{abstract}
Pedestrian crossing prediction has been a topic of active research, resulting in many new algorithmic solutions. While measuring the overall progress of those solutions over time tends to be more and more established due to the new publicly available benchmark and standardized evaluation procedures, knowing how well existing predictors react to unseen data remains an unanswered question. This evaluation is imperative as serviceable crossing behavior predictors should be set to work in various scenarii without compromising pedestrian safety due to misprediction.
To this end, we conduct a study based on direct cross-dataset evaluation. Our experiments show that current state-of-the-art pedestrian behavior predictors generalize poorly in cross-dataset evaluation scenarii, regardless of their robustness during a direct training-test set evaluation setting. In the light of what we observe, we argue that the future of pedestrian crossing prediction,\textit{ e.g.} reliable and generalizable implementations, should not be about tailoring models, trained with very little available data, and tested in a classical train-test scenario with the will to infer anything about their behavior in real life. It should be about evaluating models in a cross-dataset setting while considering their uncertainty estimates under domain shift.
\end{abstract}


%
\IEEEpeerreviewmaketitle


\section{Introduction}

\IEEEPARstart{T}{he} topic of pedestrian discrete behavior prediction is deemed essential for robust and reliable planning leading to the deployment of autonomous vehicles. While the domain has attracted significant interest in computer vision and robotics communities for the past decade, the field of research has suffered for a long time from the lack of common evaluation protocols and standardized benchmarks, making the task of comparing performance between approaches complex if not impossible to achieve. To compensate for such problems, a standardized benchmark \cite{kotseruba2021benchmark} to evaluate pedestrian behavior prediction for three datasets was recently proposed to advance research further.
While this brought a breath of fresh air to the field of pedestrian behavior prediction, we believe that current evaluation protocols do not adequately represent the applicability of existing pedestrian prediction models for real-world scenarii. Comparable studies have previously been conducted in computer vision, questioning whether recent progress on the ImageNet \cite{russakovsky2015imagenet} benchmark continues to represent meaningful generalization \cite{beyer2020we} and identifying various sources of bias and noise \cite{northcutt2021confident,stock2018convnets}. However, going beyond accuracy to evaluate a model for a high-risk application with limited amount of training data, such as pedestrian crossing prediction, has never been properly investigated. In this work, we assess how pedestrian intention prediction approaches react to small domain shifts and evaluate their generalization capability outside a standard train-test evaluation protocol. We show that all the current pedestrian behavior predictors show signs of over-fitting when evaluated during a direct training-test sets evaluation setting on those standardized benchmarks.

This problem leads to two major drawbacks for the field:
\begin{itemize}
    \item The training source being generally not dense in variety of scenarii nor in the number of examples, the results of state-of-the-art approaches on each dataset might just come from noise: this noise effect should probably be further aggravated since the existing approaches are based on deep learning, depending heavily on the quantity and quality of data where the performance of approaches scales up with the amount of training data.
    \item It prevents pedestrian behavior predictors from scaling up to real-world applications, as they are not applicable in various scenarii with small domains shifts.
\end{itemize}


\begin{figure}
 \centering
    \includegraphics[scale=0.23]{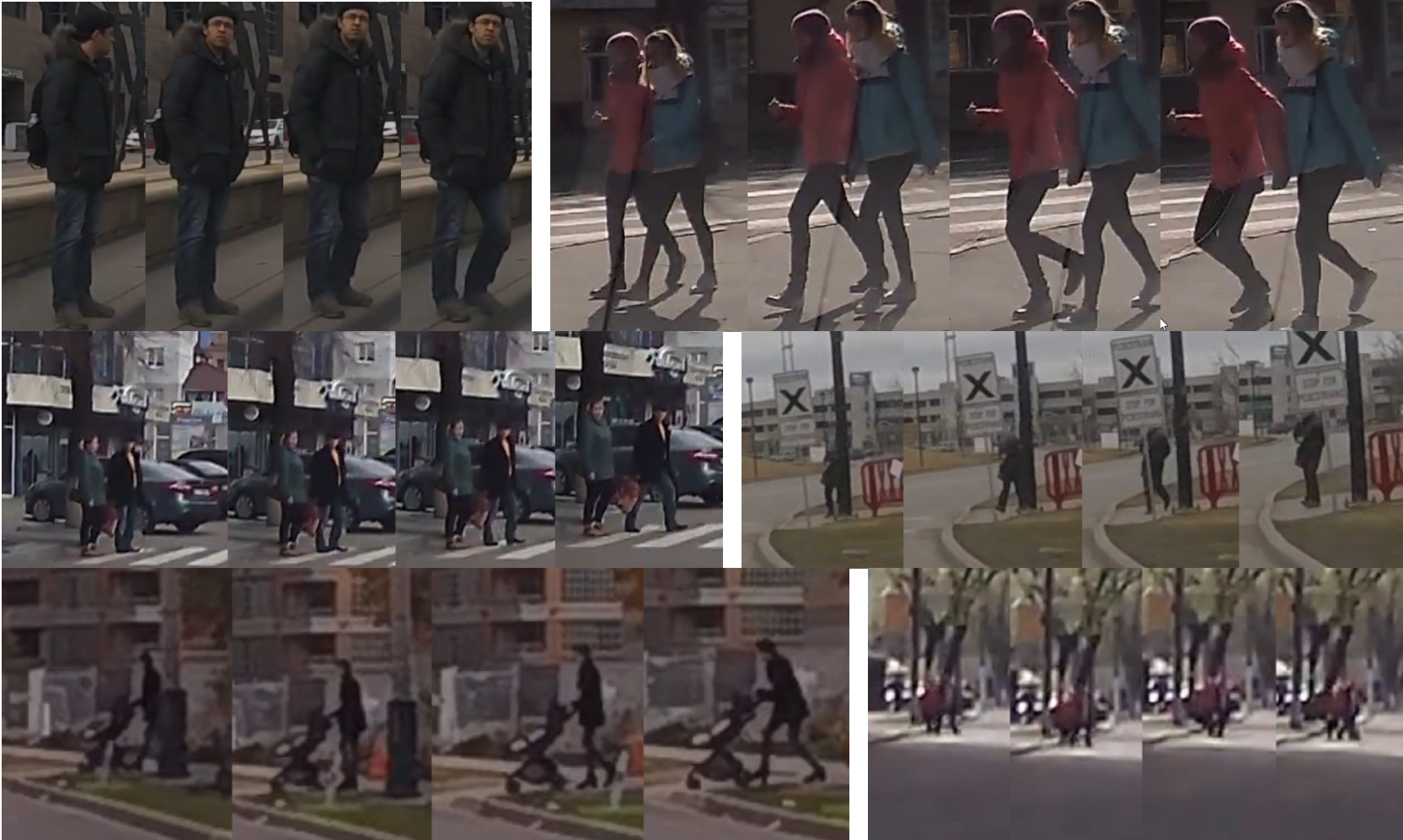}
    \caption{Examples of crossing and non-crossing pedestrians from $JAAD$ and $PIE$ datasets. The conditions under which pedestrians act from one scenario to another can differ drastically concerning input format and domain shift: pedestrian size, pedestrian positioning in the scene, illumination conditions, occlusion...}
    \label{fig:pedestrians}
\end{figure}


\begin{figure*}[!ht]
    \centerline{\includegraphics[width=0.32\textwidth]{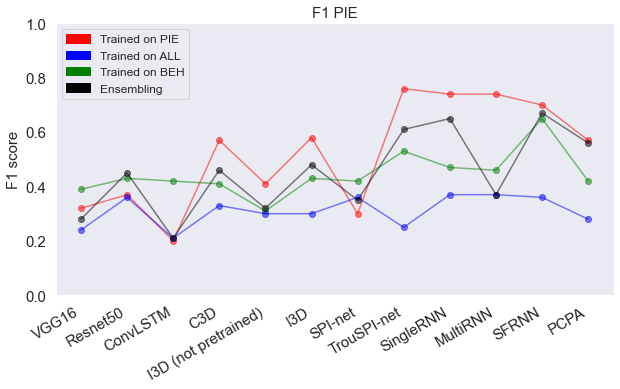}\hfill
\includegraphics[width=0.32\textwidth]{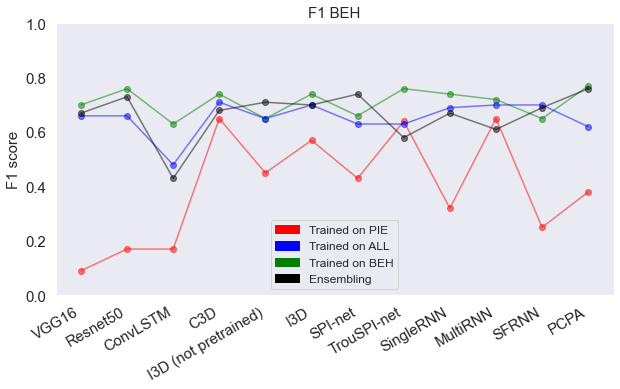}\hfill
\includegraphics[width=0.32\textwidth]{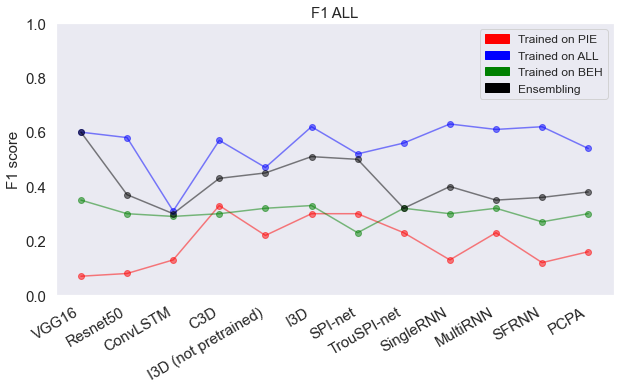}}
    \centerline{\includegraphics[width=.32\textwidth]{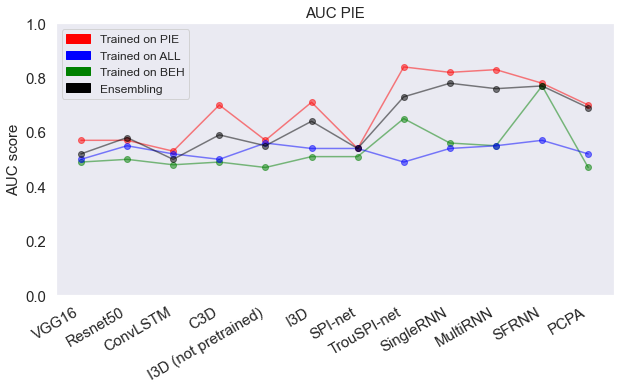}\hfill
\includegraphics[width=.32\textwidth]{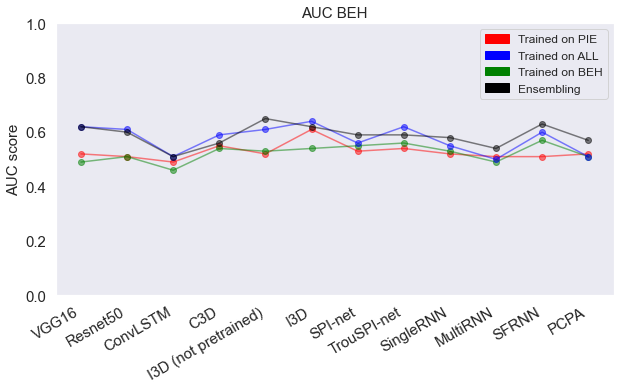}\hfill
\includegraphics[width=0.32\textwidth]{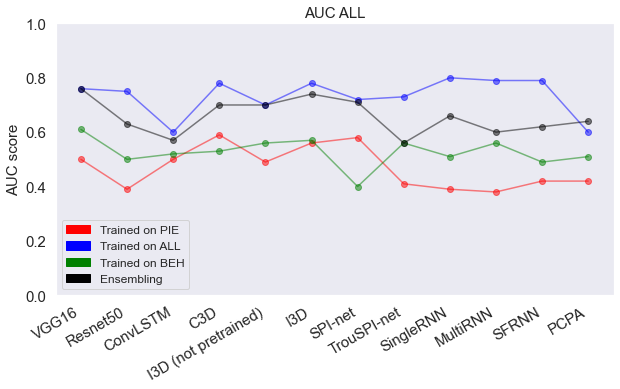}}
    \caption{Pedestrian crossing prediction performance for $PIE$, $JAAD_{behavior}$ and $JAAD_{all}$. We show a comparison between traditional single-dataset train and test evaluation on each dataset compared to cross-dataset evaluation for eleven methods representing the diversity of architectures and modalities usually used for pedestrian crossing prediction. Ensembling denotes the average prediction given by the three models trained on each dataset for one given test set.}
    \label{fig:crossdata}
\end{figure*}


The above examples recap the general motivation of this work, encouraging us to rethink the
evaluation methodology to rank current top-scoring behavior predictors from the perspective of uncertainty evaluation to small domain shifts. We argue that:
\begin{itemize}
    \item The only empirical evaluation of models in a direct train-test sets evaluation offered by the original work introducing the method is not sufficient to effectively conclude anything about its applicability in a real-world scenario. The result is often statistically non-significant during a cross-dataset evaluation scenario and leads to an ever-changing state-of-the-art.
    \item It would be more interesting to compare each method by evaluating how trustworthy are their uncertainty estimates under different domain shifts. 
\end{itemize}


\section{Related Work}

\subsection{Pedestrian Crossing Prediction}
Pedestrian crossing prediction formulates the prediction task as a binary classification problem where the objective is to determine if a pedestrian $i$ will start crossing the street given the context observed up to some time $t$. The prediction can rely on multiple sources of information, including visual features of the pedestrians and their surroundings, pedestrian kinematics, spatial positioning of the pedestrian based on 2D bounding box locations, optical flow and ego-vehicle speed. 
Early works \cite{rasouli2017ICCVW,2018arXiv181009805V}, formulated the problem as a static image classification problem with 2D Convolutions \cite{2014arXiv1409.1556S,he2016deep}, using only the last frame in the observation sequence to predict crossing behaviors. More successful approaches were designed to take into account temporal coherence in short-term motions of visual features of the pedestrians by using ConvLSTMs \cite{shi2015convolutional,8794278}, 3D Convolutions \cite{2014arXiv1412.0767T,8099985,chaabane2020looking},  or Spatio-Temporal DenseNet \cite{saleh2019real}. Approaches trying to minimize the inference time of their models by avoiding the usage of  RGB images were explored: \cite{achaji2021attention} proposes a transformer using only spatial positioning of the pedestrian based on 2D bounding box locations. Crossing prediction based on kinematics only was also explored with various available learning architectures to monitor temporal evolution of skeletal joints such as convolutions \cite{ranga2020vrunet,gesnouin2020predicting,gesnouin2021trouspi}, recurrent cells \cite{9136126,8500657} or graph-based models \cite{8917118}. 
More recently, approaches combining multiple sources of information emerged. Those approaches usually differ by the way they merge the available sources, e.g. scenes, trajectories, poses and ego-vehicle speed, and the learning architecture used to infer a crossing prediction, e.g. RNN-based models \cite{9304591,bhattacharyya2018long,2015arXiv150308909Y,kotseruba2021benchmark,yang2021predicting} or Transformer-based models \cite{lorenzo2021capformer,lorenzo2021intformer}. 


\subsection{Cross-dataset evaluation}

In its first year of existence, proposed approaches evaluated on the benchmarks \cite{kotseruba2021benchmark} constantly report higher classification scores \cite{gesnouin2021trouspi,achaji2021attention,yang2021predicting,bapu2020pedestrian,singh2021multi,zhang2021pedestrian}, giving the impression of clear improvements in pedestrian intention prediction. Usually, a new algorithm is proposed and the implicit hypothesis towards the proposed contribution is made such that it yields an improved performance over the existing state-of-the-art. To confirm such hypothesis, an empirical evaluation of the given contribution is realized in a direct train-test sets evaluation and the quality of the model is evaluated by regular classification metrics: newly proposed methods are then claimed as the new state-of-the-art as soon as they outperform previous ones even by a small margin. However, the ranking of the methods for a given task is currently only as good as the quality of the data used for comparison purposes, and the results obtained by one method on a given dataset do not always reflect its robustness in real-world applications.

\begin{figure}
\centerline{\includegraphics[width=0.5\textwidth]{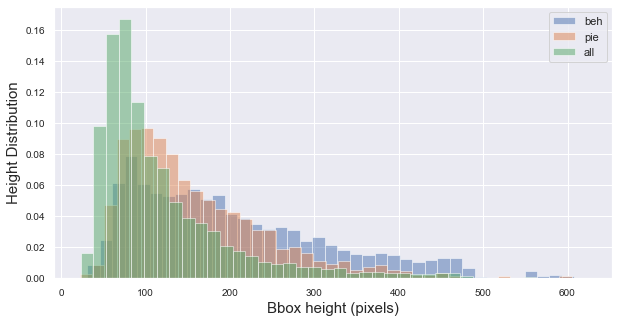}}
    \caption{Distribution of pedestrian bounding box height in pixel for $PIE$, $JAAD_{behavior}$ and $JAAD_{all}$.}
    \label{fig:bbdistrib}
\end{figure}

In this work, we evaluate how pedestrian intention prediction approaches react to small domain shifts by interchanging the training set of dataset $A$ by the training set of dataset $B$ and test it on the testing set of $A$. The given training routine is consistent across all experiments for all three datasets. This is referred throughout the paper as cross-dataset evaluation \cite{hasan2021generalizable,chen2020cdevalsumm,guo2020cross}. By adopting cross-dataset evaluation, we test the generalization abilities of several state-of-the-art pedestrian crossing predictors to distributional shift such as pedestrian size, as shown in Fig \ref{fig:bbdistrib}, pedestrian positioning in the scene, illumination conditions or occlusion as shown in Fig \ref{fig:pedestrians}. 

\subsection{Uncertainty Metrics}
In real-world scenarii, quantifying uncertainty is crucial as the input distributions are frequently shifted from the training distribution due to a number of causes such as sampling bias. Evaluating the generalization abilities of models by using cross-dataset evaluation and classification metrics only is not sufficient. In high-risk applications such as pedestrian behavior prediction, the idea that a model’s predicted probabilities of outcomes reflect true probabilities of those outcomes is mandatory for high-level decisions (\textit{i.e.}, vehicle planning module in crowded urban traffic environments). Expected Calibration Error (ECE) and Maximum Calibration Error (MCE) are standard uncertainty\footnote{Because confidence is the additive inverse of uncertainty with regard to 1, the terms are often interchanged.} metrics in this context \cite{naeini2015obtaining,guo2017calibration,heo2018uncertainty,ovadia2019can}. Predictions are divided into $M$ interval bins, we then calculate the accuracy of each bin to estimate the predicted accuracy from finite data. Let $B_{m}$ denote the set of sample indices for which prediction confidence is inside one interval bin. The accuracy of $B_{m}$ is defined as \begin{equation}
\operatorname{acc}\left(B_{m}\right)=\frac{1}{\left|B_{m}\right|} \sum_{i \in B_{m}} \mathbf{1}\left(\hat{y}_{i}=y_{i}\right)
\end{equation}

where $\hat{y}_{i}$ and $y_{i}$ are respectively the predicted and true class labels for sample $i$. The average confidence within one interval bin $B_{m}$ is defined as:
\begin{equation}
\operatorname{conf}\left(B_{m}\right)=\frac{1}{\left|B_{m}\right|} \sum_{i \in B_{m}} \hat{p}_{i}
\end{equation}

where $\hat{p}_{i}$ is the model confidence for sample $i$. Throughout our experiments, the maximum softmax probability \cite{hendrycks2016baseline} is used as the confidence score. We therefore compare each model output pseudo-probabilities to its accuracy. We obtain the following metrics to rank methods based on their calibration: 

\textbf{Expected Calibration Error} (ECE): takes a weighted average of the absolute difference in accuracy and confidence.
\begin{equation}
E C E=\sum_{m=1}^{M} \frac{\left|B_{m}\right|}{n}\left|\operatorname{acc}\left(B_{m}\right)-\operatorname{conf} \left(B_{m}\right)\right|
\end{equation}

\textbf{Maximum Calibration Error} (MCE): measures the maximum discrepancy between accuracy and confidence.
\begin{equation}
M C E=\max _{m}\left|\operatorname{acc}\left(B_{m}\right)-\operatorname{conf} \left(B_{m}\right)\right|
\end{equation}

Since the underlying binning approach has a significant impact on the accuracy and reliability of ECE and MCE, we use an adaptive binning strategy \cite{ding2020revisiting} instead of a uniform partition\footnote{https://github.com/yding5/AdaptiveBinning}.





\section{Generalization Capabilities}
\subsection{Datasets and Implementation Details}
For this evaluation, we use two large public naturalistic datasets for studying pedestrian behavior prediction: $JAAD$ \cite{Rasouli_2017_ICCV} and $PIE$ \cite{rasouli2017they}. These datasets are typically obtained by a vehicle-mounted camera as it navigates through crowded urban traffic environments: $JAAD$ contains 346 clips and focuses on pedestrians intending to cross, $PIE$ contains 6 hours of continuous footage and provides annotations for all pedestrians sufficiently close to the road regardless of their intent to cross in front of the ego-vehicle and provides more diverse behaviors of pedestrians. The $JAAD$ dataset is split into $JAAD_{behavior}$ and $JAAD_{all}$. $JAAD_{behavior}$ is biased towards pedestrians attempting to cross the street (402 crossing out of 648) and the smallest dataset available. $JAAD_{all}$ adds all visible pedestrians in $JAAD$, regardless of their position in the scene and contains more non-crossing pedestrians (490 crossing out of 2580). Similarly, $PIE$ contains more non-crossing pedestrians (512 crossing out of 1842). All three datasets are heavily skewed towards one class. To compensate for such significant datasets shifts label-wise, we train all our models using class weights inversely proportional to the percentage of samples for each class. Following the existing evaluation procedures \cite{kotseruba2021benchmark}, we use the same data sampling method, the same splits and the same inputs sets for our experiments\footnote{https://github.com/ykotseruba/PedestrianActionBenchmark}. However, we disregard the ego-vehicle speed input for all our models as the sensor data used for the ego-vehicle speed is only available for $PIE$ and could not be used for cross-dataset evaluation purposes. The observation length for all models is fixed at 16 frames. The sample overlap is set to 0.8 for both $PIE$ and $JAAD$.
We report the results using standard binary classification metrics: AUC and F1 Score and standard confidence calibration metrics: adaptive ECE and MCE.

\subsection{Baselines and state-of-the-art models}
We select a subset of methods from the pedestrian crossing prediction literature, and more broadly, action recognition literature for their prevalence, practical applicability and diversity in terms of architectures and input modalities. These include:

\begin{itemize}
    \item \textbf{VGG16} \cite{2014arXiv1409.1556S} and \textbf{Resnet50} \cite{he2016deep} : two baseline static models that use only the last frame in the observation sequence to predict the crossing behavior of a pedestrian.
    
        \begin{figure}[]
 \centering
    \includegraphics[scale=0.4]{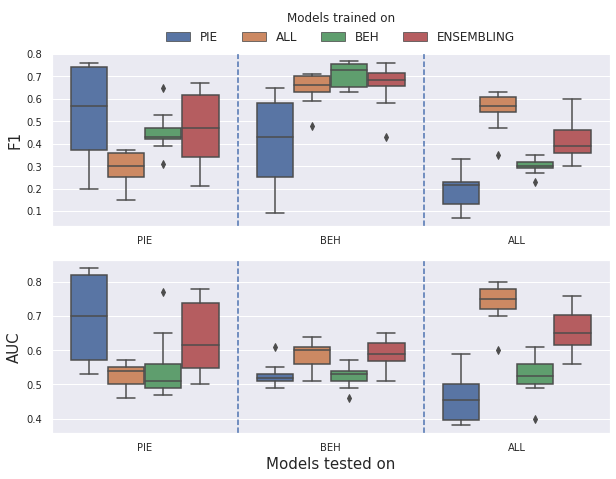}
    \caption{Distribution of the performance of the eleven selected approaches when evaluated in a direct train-test scenario and when evaluated under cross-dataset scenario.}
    \label{fig:boxplots}
\end{figure}

    \item \textbf{ConvLSTM} \cite{shi2015convolutional}: A model using a stack of images as input, pre-process those images with pre-trained CNN and apply ConvLSTM on those features.
    \item \textbf{Convolutional-3D} (\textit{C3D}) \cite{2014arXiv1412.0767T} and \textbf{Inflated-3D} (\textit{I3D}) \cite{8099985}: two models pretrained on Sports1M \cite{karpathy2014large} using a stack of images as input and applying 3D convolutions to extract features.
    
    \item \textbf{SPI-net} \cite{gesnouin2020predicting} and \textbf{TrouSPI-net} \cite{gesnouin2021trouspi}: two multi-modal models relying on pedestrians' pose kinematics extracted by OpenPose \cite{cao2017realtime}, relative euclidean distance of key-points and evolution of the pedestrian spatial positioning. Poses sequences are converted into 2D image-like spatio-temporal representations and self-spatio-temporal attention is applied via CNN-based models for multiple time resolutions. Each remaining feature is independently processed via either U-GRUs \cite{rozenberg2021asymmetrical} or feed forward neural network and fused by either applying temporal and modality attention or sent to a \textit{fc} layer to predict crossing behaviors.

    \item \textbf{SingleRNN} \cite{9304591}, \textbf{Multi-stream RNN} (\textit{MultiRNN}) \cite{bhattacharyya2018long} and \textbf{Stacked with multilevel Fusion RNN} (\textit{SFRNN}) \cite{Rasouli2019PedestrianAA}: Three multi-modal models relying on RGB Images extracted by VGG16 \cite{2014arXiv1409.1556S}, pose kinematics extracted by OpenPose \cite{cao2017realtime} and evolution of the pedestrian spatial positioning. Input features are either  concatenated into a single vector and sent to a recurrent network followed by a \textit{fc} layer for crossing prediction, either processed independently by GRUs \cite{chung2014empirical} and the hidden state of GRUs are then concatenated and sent into a \textit{fc} layer for crossing prediction or either processed by GRUs \cite{chung2014empirical} and fused gradually at different levels of processing and complexity.
    \item \textbf{Pedestrian Crossing Prediction with Attention} (\textit{PCPA}) \cite{kotseruba2021benchmark}: A multi-modal model relying on RGB images extracted by C3D \cite{2014arXiv1412.0767T}, pose kinematics extracted by OpenPose \cite{cao2017realtime} and evolution of the pedestrian spatial positioning. Non-images features are independently encoded by GRUs \cite{chung2014empirical} and each is fed to a temporal attention block. 3D Convoluted features are flattened and fed into a \textit{fc} layer. Modality attention is then applied to all the branches to fuse them into a single representation by weighted summation of the information from individual modalities. 
\end{itemize}

\begin{table*}
\centering
\scalebox{0.9}{
\begin{tabular}{l  a  b  a |  b  a  b |  a  b  a   | b   a  b } 
\toprule
\begin{tabular}[c]{@{}l@{}}\\\end{tabular} & \multicolumn{3}{l}{\textbf{AUC ($\uparrow$)}} & \multicolumn{3}{l}{\textbf{F1($\uparrow$)}} & \multicolumn{3}{l}{\textbf{ECE($\downarrow$) }} & \multicolumn{3}{l}{\textbf{MCE ($\downarrow$) }}  \\ 
\hline
\textbf{Method}                            & \textbf{pie \textcolor{darkgray}{\tiny{($\pm$0.02})}}  & \textbf{beh \textcolor{darkgray}{\tiny{($\pm$0.02})}}  & \textbf{all \textcolor{darkgray}{\tiny{($\pm$0.01})}}                & \textbf{pie \textcolor{darkgray}{\tiny{($\pm$0.03})}}  & \textbf{beh \textcolor{darkgray}{\tiny{($\pm$0.01})}}  & \textbf{all \textcolor{darkgray}{\tiny{($\pm$0.02})}}                & \textbf{pie \textcolor{darkgray}{\tiny{($\pm$0.01})}}  & \textbf{beh \textcolor{darkgray}{\tiny{($\pm$0.02})}}   & \textbf{all \textcolor{darkgray}{\tiny{($\pm$0.02})}}               & \textbf{pie \textcolor{darkgray}{\tiny{($\pm$0.02})}}   & \textbf{beh \textcolor{darkgray}{\tiny{($\pm$0.03})}}   & \textbf{all \textcolor{darkgray}{\tiny{($\pm$0.03})}}                \\ 
\hline
VGG16 \cite{2014arXiv1409.1556S}                                    & 0.52 & \textbf{0.62} & \textbf{0.76}               & 0.28 & 0.67 & \textbf{0.60}               & 0.07 & 0.06 & 0.20             & 0.24 & \textbf{0.13} & 0.25              \\
Resnet \cite{he2016deep}                                    & 0.58 & 0.60 & 0.63               & 0.45 & 0.68 & 0.37               & 0.09 & 0.05 & \textbf{0.04}             & 0.37 & \textbf{0.15} & 0.44              \\ 
\hdashline
ConvLSTM \cite{shi2015convolutional}                                  & 0.50 & 0.51 & 0.57               & 0.21 & 0.43 & 0.30               & 0.09 & 0.14 & 0.10             & 0.22 & 0.25 & 0.41              \\
C3D \cite{2014arXiv1412.0767T}                                        & 0.59 & 0.56 & 0.70               & 0.46 & 0.73 & 0.43               & 0.17 & 0.08 & \textbf{0.03}             & 0.43 & \textbf{0.12} & \textbf{0.11}              \\
I3D \cite{8099985}                                       & 0.64 & \textbf{0.62} & \textbf{0.74}               & 0.48 & 0.71 & 0.51               & \textbf{0.05} & 0.08 & 0.09             & \textbf{0.13} & \textbf{0.15} & \textbf{0.16}              \\ 
\hdashline
PCPA  \cite{kotseruba2021benchmark}                                     & 0.69 & 0.57 & 0.64               & 0.56 & 0.67 & 0.38               & 0.12 & 0.04 & 0.12             & 0.36 & \textbf{0.13} & 0.28              \\
SingleRNN \cite{9304591}                                 & \textbf{0.78} & 0.58 & 0.66               & \textbf{0.65} & 0.69 & 0.40               & 0.09 & \textbf{0.02} & 0.09             & 0.16 & \textbf{0.15} & \textbf{0.14 }             \\
MultiRNN  \cite{bhattacharyya2018long}                                 & \textbf{0.76} & 0.54 & 0.60               & \textbf{0.64} & \textbf{0.74} & 0.35               & \textbf{0.06} & 0.08 & 0.19             & \textbf{0.13} & \textbf{0.17} & 0.378              \\
SFRNN \cite{Rasouli2019PedestrianAA}                                      & \textbf{0.77} & \textbf{0.63} & 0.62               & \textbf{0.67} & 0.58 & 0.36               & 0.07 & 0.08 & 0.11             & \textbf{0.11} & 0.29 & \textbf{0.16}              \\ 
\hdashline
Spi-Net \cite{gesnouin2020predicting}                                   & 0.54 & 0.59 & 0.71               & 0.35 & 0.61 & 0.50               & 0.10 & 0.07 & 0.22             & 0.30 & \textbf{0.15} & 0.33              \\
TrouSPI-net \cite{gesnouin2021trouspi}                               & 0.73 & 0.59 & 0.56               & 0.61 & \textbf{0.76} & 0.32               & 0.07 & 0.05 & 0.24             & \textbf{0.12} & \textbf{0.13} & 0.41              \\
\bottomrule
\end{tabular}}
\caption{Average prediction given by the three models trained on each training sets for one given test-set (Ensembling). In addition to classification metrics (we use arrows to indicate which direction is better), we compare models with predictive uncertainty metrics such as Expected Calibration Error (ECE) and Maximum Calibration Error (MCE). Dashed lines separate different types of architectures}
\label{table:resens}
\end{table*}

    \begin{figure}[]
 \centering
    \includegraphics[scale=0.32]{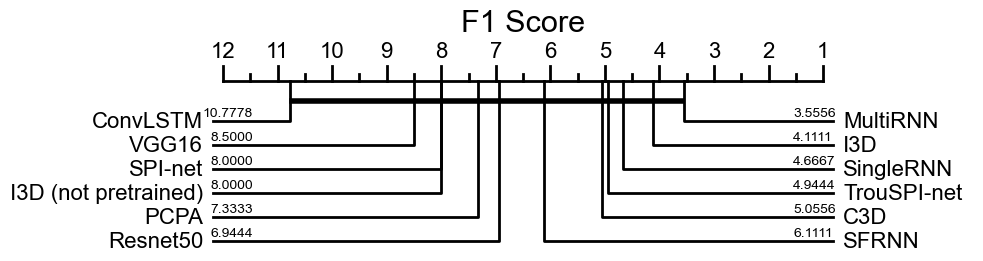}
    \includegraphics[scale=0.32]{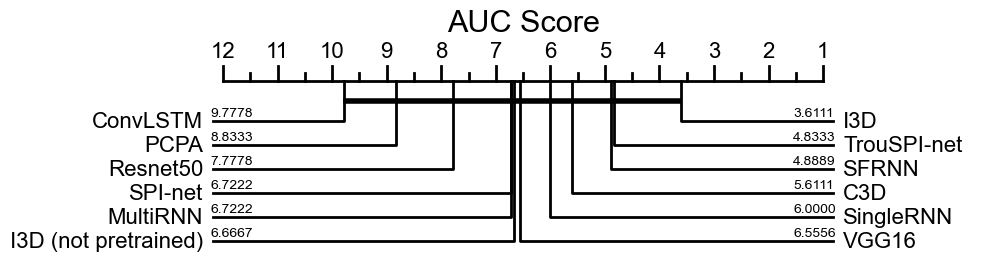}
    \caption{Critical Difference Diagram \cite{demvsar2006statistical}: first a Friedman test is performed to reject the null hypothesis, we then proceed with a post-hoc analysis based on the Wilcoxon-Holm method. We compare the robustness of classifiers over multiple training and testing sets shifts.
    We can see how each method ranks on average. A thick horizontal line groups a set of classifiers that are not significantly different ($\alpha = 0.1$).
}
    \label{fig:criticaldiffdiagram}
\end{figure}

\subsection{Cross-dataset Evaluation Results}
We present the coarse results of our cross-dataset evaluation on Fig \ref{fig:crossdata}. For readability purposes, the corresponding critical difference diagram is reported on Fig \ref{fig:criticaldiffdiagram} and the average distribution of performance of the selected approaches is reported on Fig \ref{fig:boxplots}. The results of the average prediction given by the three models trained on each training set for one given test set is reported on Table \ref{table:resens}. 

As expected, all methods, regardless of their architecture or input modalities, suffer a consequent performance drop when trained on $PIE$ and tested on $JAAD$ and vice versa. Fig \ref{fig:boxplots} shows that however robust the individual classifier is, there is a general trend for classifiers to decline when exposed to a different test set than the expected one. This is consistent towards all our experiments with the exception of $JAAD_{behavior}$. $JAAD_{all}$ being an extension to the set of samples with behavioral annotations, $JAAD_{all}$ "generalizes" well on $JAAD_{behavior}$ but unsurprisingly, the converse is far from true. Even when trained on a relatively diverse dataset ($PIE$) and inferred on a smaller one in comparison ($JAAD_{behavior}$), selected methods barely show signs of generalization.
More alarming, some methods even under-performed a random binary guess based on class distribution when exposed to a different testing set than the expected one. While the task, standardized inputs and observation length are the same across all three datasets, none of the tested models reaches a satisfactory level of generalization across any other testing set.

When it comes to compare performance towards small domain-shift at the granular level of individuals methods, the critical difference diagram reported on Fig \ref{fig:criticaldiffdiagram}, shows that none of the selected methods arise as a clear winner when it comes to cross-dataset ranking. More importantly, the obtained ranks of each method when evaluated under cross-dataset evaluation is far from the one we usually consider when developing pedestrian crossing behavior predictors:  some general methods such as I3D or C3D are on par with multi-modal methods specifically designed to tackle the problem of pedestrian crossing prediction. This confirms the importance of rethinking the evaluation methodology of our approaches.

The ensembling provided in Table \ref{table:resens}, is the closest plausible approximation of the selected models' robustness for real-world application as it integrates all available conditions and training instances while removing the sampling biases of each specific training set. It shows that the only empirical evaluation of models in a direct train-test sets evaluation is not sufficient to effectively conclude anything about its applicability in a real-world scenario. This also demonstrates that the use of classification metrics alone is not representative of the overall capacity of the models. For two given models which are equivalent with respect to classification metrics (AUC or F1 score), their calibration (ECE and MCE) can differ drastically. This supports our argument that the usage of uncertainty metrics should complement the metrics conventionally used in order to obtain a comprehensive view of the robustness of existing approaches.

\begin{figure}[]
 \centering
    \includegraphics[scale=0.25]{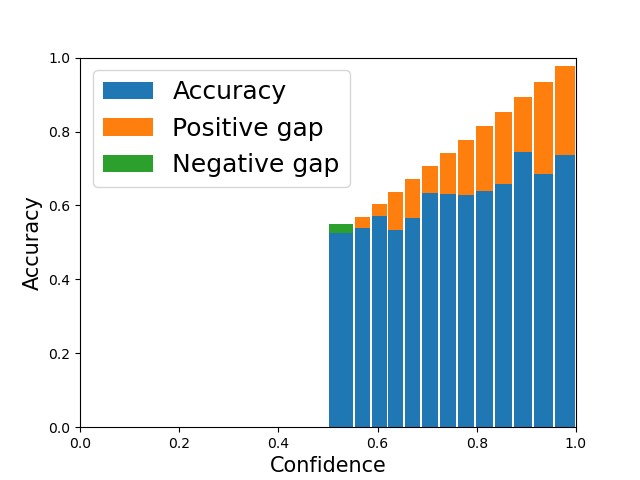}
    \includegraphics[scale=0.25]{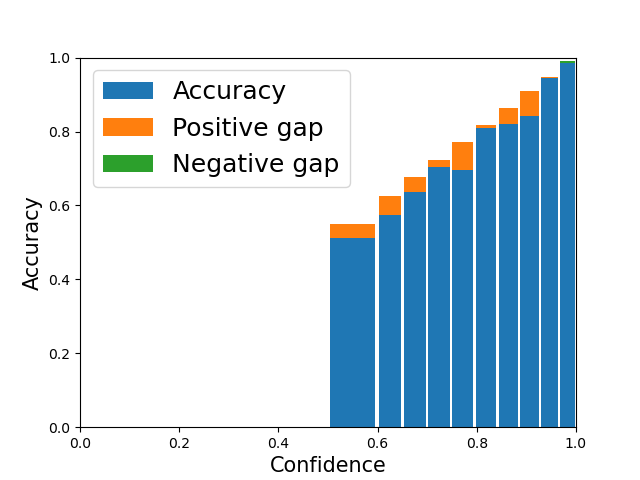}\hfill
    \includegraphics[scale=0.25]{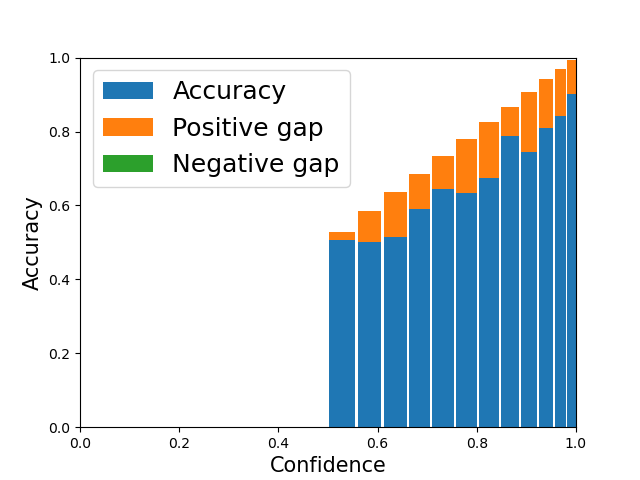}
    \includegraphics[scale=0.25]{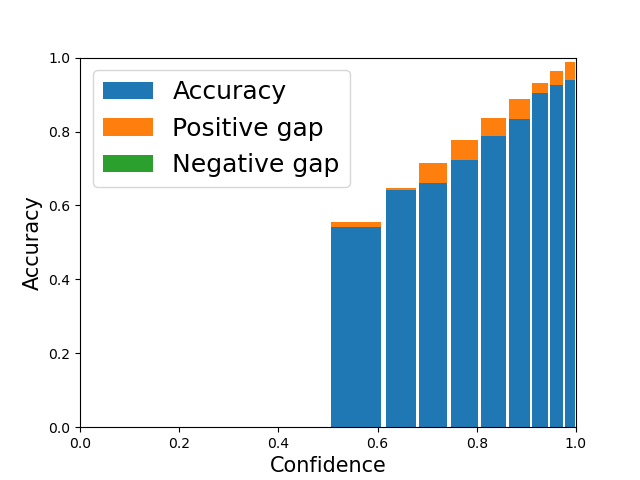}\hfill
    \includegraphics[scale=0.25]{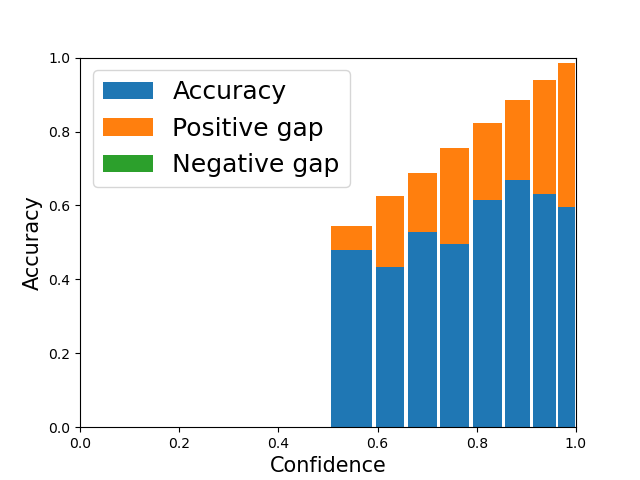}
    \includegraphics[scale=0.25]{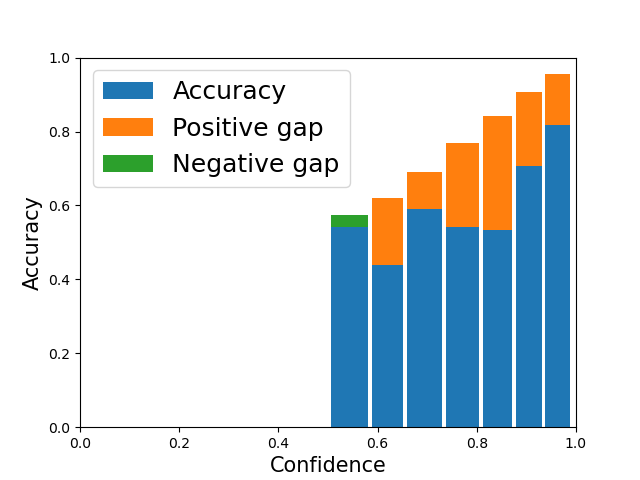}\hfill
    \caption{Reliability Diagrams between I3D \cite{8099985} randomly initialized (left) and pre-trained on Sports1M \cite{karpathy2014large}(right) on $PIE$, $JAAD_{all}$ and $JAAD_{behavior}$ datasets. If the model is perfectly calibrated, then the diagram plots the identity function. Any deviation from a perfect diagonal represents miscalibration: the model is either overconfident (orange) or subconfident (green).}
    \label{fig:reliabilityDiagram}
\end{figure}

\begin{table*}[h]
\centering
\scalebox{0.75}{
\begin{tabular}{l  a  b  a |  b  a  b |  a  b  a   | b   a  b } 
\toprule
\begin{tabular}[c]{@{}l@{}}\\\end{tabular} & \multicolumn{3}{l}{\textbf{AUC} ($\uparrow$)}                & \multicolumn{3}{l}{\textbf{F1} ($\uparrow$)}               & \multicolumn{3}{l}{\textbf{ECE} ($\downarrow$)}                   & \multicolumn{3}{l}{\textbf{MCE} ($\downarrow$)}                  \\ 
\hline
\textbf{Method}                            & \textbf{pie}            & \textbf{beh}           & \textbf{all}            & \textbf{pie }          & \textbf{beh}            & \textbf{all}           & \textbf{pie}             & \textbf{beh}             & \textbf{all}             & \textbf{pie}            & \textbf{beh}             & \textbf{all}             \\ 
\hline
Non-pretrained                             & 0.55~\textcolor{darkgray}{\tiny{($\pm$0.04)}}   & 0.50~\textcolor{darkgray}{\tiny{($\pm$0.02)}}  & 0.69~\textcolor{darkgray}{\tiny{($\pm$ 0.05)}}  & 0.34~\textcolor{darkgray}{\tiny{($\pm$0.09)}}  & 0.65~\textcolor{darkgray}{\tiny{($\pm$0.08)}}   & 0.54~\textcolor{darkgray}{\tiny{($\pm$0.06)}}  & 0.205~\textcolor{darkgray}{\tiny{($\pm$0.062)}}  & 0.184~\textcolor{darkgray}{\tiny{($\pm$ 0.089)}} & 0.111~\textcolor{darkgray}{\tiny{($\pm$ 0.069)}} & 0.290~\textcolor{darkgray}{\tiny{($\pm$0.039)}} & 0.338~\textcolor{darkgray}{\tiny{($\pm$0.191)}}  & 0.162~\textcolor{darkgray}{\tiny{($\pm$0.075)}}  \\
Ens-Nonpretrained                          & 0.59~\textcolor{darkgray}{\tiny{($\pm$0.06)}}   & 0.58~\textcolor{darkgray}{\tiny{($\pm$0.05)}}  & 0.64~\textcolor{darkgray}{\tiny{($\pm$0.04)}}   & 0.37~\textcolor{darkgray}{\tiny{($\pm$0.10)}}  & 0.55~\textcolor{darkgray}{\tiny{($\pm$0.10)}}   & 0.39~\textcolor{darkgray}{\tiny{($\pm$0.05)}}  & 0.065~\textcolor{darkgray}{\tiny{($\pm$0.022)}}  & 0.131~\textcolor{darkgray}{\tiny{($\pm$0.058)}}  & 0.091~\textcolor{darkgray}{\tiny{($\pm$0.026)}}  & 0.248~\textcolor{darkgray}{\tiny{($\pm$0.122)}} & 0.644~\textcolor{darkgray}{\tiny{($\pm$0.152)}}  & 0.280~\textcolor{darkgray}{\tiny{($\pm$0.172)}}  \\ 
\hdashline
Deterministic                              & 0.72~\textcolor{darkgray}{\tiny{($\pm$0.01)}}   & 0.56~\textcolor{darkgray}{\tiny{($\pm$0.03)}}  & 0.76~\textcolor{darkgray}{\tiny{($\pm$0.03)}}   & 0.60~\textcolor{darkgray}{\tiny{($\pm$0.02)}}  & 0.74~\textcolor{darkgray}{\tiny{($\pm$0.02)}}   & 0.61~\textcolor{darkgray}{\tiny{($\pm$0.02)}}  & 0.026~\textcolor{darkgray}{\tiny{($\pm$0.007)}}  & 0.143~\textcolor{darkgray}{\tiny{($\pm$ 0.020)}} & 0.054~\textcolor{darkgray}{\tiny{($\pm$0.010)}}  & 0.063~\textcolor{darkgray}{\tiny{($\pm$0.007)}} & 0.239~\textcolor{darkgray}{\tiny{($\pm$ 0.030)}} & 0.118~\textcolor{darkgray}{\tiny{($\pm$0.020)}}  \\
Ens-Deterministic                          & 0.64~\textcolor{darkgray}{\tiny{($\pm$0.01)}}   & 0.62~\textcolor{darkgray}{\tiny{($\pm$0.01)}}  & 0.73~\textcolor{darkgray}{\tiny{($\pm$ 0.1)}}   & 0.49~\textcolor{darkgray}{\tiny{($\pm$0.02)}}  & 0.70~\textcolor{darkgray}{\tiny{($\pm$0.01)}}   & 0.51~\textcolor{darkgray}{\tiny{($\pm$0.01)}}  & 0.053~\textcolor{darkgray}{\tiny{($\pm$0.003)}}  & 0.080~\textcolor{darkgray}{\tiny{($\pm$0.001)}}  & 0.097~\textcolor{darkgray}{\tiny{($\pm$0.016)}}  & 0.120~\textcolor{darkgray}{\tiny{($\pm$0.013)}} & 0.138~\textcolor{darkgray}{\tiny{($\pm$0.024)}}  & 0.172~\textcolor{darkgray}{\tiny{($\pm$0.022)}}  \\ 
\hdashline
MC Dropout                                 & 0.73~\textcolor{darkgray}{\tiny{($\pm$0.01)}}   & 0.55~\textcolor{darkgray}{\tiny{($\pm$0.01)}}  & \textbf{0.78}~\textcolor{darkgray}{\tiny{($\pm$0.01)}}   & 0.61~\textcolor{darkgray}{\tiny{($\pm$0.01)}}  & 0.67~\textcolor{darkgray}{\tiny{($\pm$0.01)}}   & 0.60~\textcolor{darkgray}{\tiny{($\pm$ 0.01)}} & 0.064~\textcolor{darkgray}{\tiny{($\pm$ 0.003)}} & \textbf{0.063}~\textcolor{darkgray}{\tiny{($\pm$0.005)}}  & 0.040~\textcolor{darkgray}{\tiny{($\pm$0.002)}}  & 0.106~\textcolor{darkgray}{\tiny{($\pm$0.004)}} & \textbf{0.134}~\textcolor{darkgray}{\tiny{($\pm$0.012)}}  & \textbf{0.059}~\textcolor{darkgray}{\tiny{($\pm$0.009)}}  \\
Ens-MC Dropout                             & 0.61~\textcolor{darkgray}{\tiny{($\pm$0.01)}}   & 0.61~\textcolor{darkgray}{\tiny{($\pm$0.01)}}  & 0.73~\textcolor{darkgray}{\tiny{($\pm$0.01)}}   & 0.42~\textcolor{darkgray}{\tiny{($\pm$0.01)}}  & 0.49~\textcolor{darkgray}{\tiny{($\pm$0.02)}}   & 0.53~\textcolor{darkgray}{\tiny{($\pm$0.01)}}  & 0.053~\textcolor{darkgray}{\tiny{($\pm$0.002)}}  & 0.053~\textcolor{darkgray}{\tiny{($\pm$0.003)}}  & 0.129~\textcolor{darkgray}{\tiny{($\pm$0.002)}}  & 0.096~\textcolor{darkgray}{\tiny{($\pm$0.005)}} & 0.120~\textcolor{darkgray}{\tiny{($\pm$0.013)}}  & 0.181~\textcolor{darkgray}{\tiny{($\pm$0.007)}}  \\ 
\hdashline
TempScaling                                & 0.70~\textcolor{darkgray}{\tiny{($\pm$0.02)}}   & \textbf{0.58}~\textcolor{darkgray}{\tiny{($\pm$0.02)}}  & 0.76~\textcolor{darkgray}{\tiny{($\pm$0.01)}}   & 0.57~\textcolor{darkgray}{\tiny{($\pm$0.03)}}  & 0.72~\textcolor{darkgray}{\tiny{($\pm$0.01)}}   & 0.61~\textcolor{darkgray}{\tiny{($\pm$ 0.02)}} & \textbf{0.020}~\textcolor{darkgray}{\tiny{($\pm$0.005)}}  & 0.070 \textcolor{darkgray}{\tiny{($\pm$ 0.020)}} & 0.037~\textcolor{darkgray}{\tiny{($\pm$0.08)}}   & \textbf{0.050}~\textcolor{darkgray}{\tiny{($\pm$0.010)}} & 0.300~\textcolor{darkgray}{\tiny{($\pm$0.130)}}  & 0.146~\textcolor{darkgray}{\tiny{($\pm$0.035)}}  \\
Ens-TempScaling                            & 0.61~\textcolor{darkgray}{\tiny{($\pm$ 0.01)}}  & \textbf{0.66}~\textcolor{darkgray}{\tiny{($\pm$0.01)}}  & \textbf{0.75}~\textcolor{darkgray}{\tiny{($\pm$0.01)}}   & 0.41~\textcolor{darkgray}{\tiny{($\pm$0.02)}}  & 0.69~\textcolor{darkgray}{\tiny{($\pm$0.01)}}   & \textbf{0.56}~\textcolor{darkgray}{\tiny{($\pm$0.01)}}  & 0.058~\textcolor{darkgray}{\tiny{($\pm$0.004)}}  & 0.054~\textcolor{darkgray}{\tiny{($\pm$0.008)}}  & 0.142~\textcolor{darkgray}{\tiny{($\pm$0.015)}}  & 0.127~\textcolor{darkgray}{\tiny{($\pm$0.012)}} & 0.237~\textcolor{darkgray}{\tiny{($\pm$0.068)}}   & 0.215~\textcolor{darkgray}{\tiny{($\pm$0.016)}}  \\ 
\hdashline
LL Dropout                                 & 0.71~\textcolor{darkgray}{\tiny{($\pm$0.01)}}   & 0.54~\textcolor{darkgray}{\tiny{($\pm$0.003)}} & \textbf{0.78}~\textcolor{darkgray}{\tiny{($\pm$0.001)}}  & 0.59~\textcolor{darkgray}{\tiny{($\pm$0.01)}}  & 0.74~\textcolor{darkgray}{\tiny{($\pm$ 0.001)}} & \textbf{0.62}~\textcolor{darkgray}{\tiny{($\pm$0.003)}} & \textbf{0.020} \textcolor{darkgray}{\tiny{($\pm$ 0.003)}}  & 0.155~\textcolor{darkgray}{\tiny{($\pm$0.003)}}  & 0.061~\textcolor{darkgray}{\tiny{($\pm$ 0.001)}} & 0.063~\textcolor{darkgray}{\tiny{($\pm$0.012)}} & 0.247~\textcolor{darkgray}{\tiny{($\pm$0.01)}}   & 0.107~\textcolor{darkgray}{\tiny{($\pm$0.016)}}  \\
Ens-LL Dropout                             & 0.65~\textcolor{darkgray}{\tiny{($\pm$ 0.002)}} & 0.62~\textcolor{darkgray}{\tiny{($\pm$0.003)}} & 0.73~\textcolor{darkgray}{\tiny{($\pm$0.002)}}  & 0.49~\textcolor{darkgray}{\tiny{($\pm$0.003)}} & 0.70~\textcolor{darkgray}{\tiny{($\pm$0.001)}}  & 0.50~\textcolor{darkgray}{\tiny{($\pm$0.003)}} & 0.052~\textcolor{darkgray}{\tiny{($\pm$0.003)}}  & 0.081~\textcolor{darkgray}{\tiny{($\pm$0.002)}}  & 0.098~\textcolor{darkgray}{\tiny{($\pm$0.002)}}  & \textbf{0.105}~\textcolor{darkgray}{\tiny{($\pm$0.002)}} & 0.145~\textcolor{darkgray}{\tiny{($\pm$0.023)}}  & 0.176~\textcolor{darkgray}{\tiny{($\pm$0.005)}}  \\ 
\hdashline
LL SVI                                     & \textbf{0.74}~\textcolor{darkgray}{\tiny{($\pm$0.01)}}   & 0.53~\textcolor{darkgray}{\tiny{($\pm$0.01)}}  & 0.77~\textcolor{darkgray}{\tiny{($\pm$ 0.003)}} & \textbf{0.62}~\textcolor{darkgray}{\tiny{($\pm$0.01)}}  & \textbf{0.76}~\textcolor{darkgray}{\tiny{($\pm$0.01)}}   & 0.57~\textcolor{darkgray}{\tiny{($\pm$0.004)}} & \textbf{0.021}~\textcolor{darkgray}{\tiny{($\pm$0.002)}}  & 0.162~\textcolor{darkgray}{\tiny{($\pm$0.004)}}  & \textbf{0.026}~\textcolor{darkgray}{\tiny{($\pm$0.003)}}  & 0.059~\textcolor{darkgray}{\tiny{($\pm$0.009)}} & 0.214~\textcolor{darkgray}{\tiny{($\pm$0.006)}}  & \textbf{0.054}~\textcolor{darkgray}{\tiny{($\pm$0.009)}}   \\
Ens-LL SVI                                 & \textbf{0.68}~\textcolor{darkgray}{\tiny{($\pm$0.003)}}  & 0.61~\textcolor{darkgray}{\tiny{($\pm$0.003)}} & 0.69~\textcolor{darkgray}{\tiny{($\pm$0.003)}}  & \textbf{0.55}~\textcolor{darkgray}{\tiny{($\pm$0.003)}} & \textbf{0.73}~\textcolor{darkgray}{\tiny{($\pm$0.002)}}  & 0.43~\textcolor{darkgray}{\tiny{($\pm$0.004)}} & \textbf{0.045}~\textcolor{darkgray}{\tiny{($\pm$0.003)}}  & \textbf{0.036}~\textcolor{darkgray}{\tiny{($\pm$0.003)}}  & \textbf{0.046}~\textcolor{darkgray}{\tiny{($\pm$0.005)}}  & 0.146~\textcolor{darkgray}{\tiny{($\pm$0.027)}} & \textbf{0.075}~\textcolor{darkgray}{\tiny{($\pm$0.009)}}  & \textbf{0.133}~\textcolor{darkgray}{\tiny{($\pm$0.016)}}  \\
\bottomrule

\end{tabular}}
\caption{Average Pedestrian Crossing Prediction performance for $PIE$, $JAAD_{behavior}$ and $JAAD_{all}$ (5 runs). Dashed lines separate each probabilistic deep learning baseline. Each baseline is tested twice: first, in a classical train-test evaluation protocol and then tested by ensembling all three models trained on each training set to evaluate its robustness to small domain shift. We highlight the highest scores for each metric and for both evaluation protocols: train-test or ensembling.}
\label{tab:final}

\end{table*}

\subsection{Role of pre-training in uncertainty calibration}

Table \ref{table:resens} illustrates that generic baseline methods (\textit{i.e.} VGG16, C3D, I3D) pre-trained on well diverse and dense datasets further away from the target domain, benefit in terms of generalization and uncertainty calibration as they are on par with the methods specifically designed to tackle the problem of pedestrian crossing prediction, which was not the case in a simple train-test evaluation setting.

To better isolate the effects of pre-training with larger datasets we consider two I3D \cite{8099985} but trained with different configuration: the first one being randomly initialized and the second one being pre-trained on Sports1M \cite{karpathy2014large}. We assess their performance on the same datasets and report our findings in Fig \ref{fig:reliabilityDiagram}. We show that pre-trained models significantly outperform randomly initialized models across all three datasets in terms of calibration. 
As far as robustness aspects towards small domain shifts are concerned, this may become an important factor to consider when designing pedestrian crossing behavior approaches for real-world scenarii. The training source being generally not dense in variety of conditions nor in the number of examples, the results provided on each dataset might just come from noise or over-fitting models on testing sets. Pre-training well-established models on diverse and dense datasets further away from the target domain before fine-tunning to our target task might prove efficient and mandatory for the next step of pedestrian crossing behavior prediction: generalization and vehicle implementation.




\section{Improving Uncertainty Calibration}

For the very same approach, there is a significant discrepancy between traditional train-test and cross-dataset evaluation results. This calls into question the reliability of current methods in regard to their capacity to generalize. In addition, we have shown that the standard classification metrics are not sufficient to reliably evaluate an approach since the use of uncertainty metrics raises additional issues that are not reflected otherwise. We are confident that the future breakthroughs in the area will not occur by outperforming current state-of-the-art by a small margin on conventionally used evaluation protocols as they currently fail to provide the big picture of pedestrian crossing behavior prediction. 

As we encourage the community to change the direction in which we are taking the research field, we investigate how additional baselines from the probabilistic deep learning literature improve the generalization ability of pedestrian behavior predictors towards small domain shifts. We believe that those methods could prove useful for the next generation of predictors and present our results with the intention that they will serve as a baseline for future work addressing our prescriptions. 

\subsection{Baselines from the probabilistic deep learning literature}
Below, we present the  selected methods from the probabilistic deep learning literature applied on top of an I3D \cite{8099985} model:
\begin{itemize}
    \item \textbf{Non-pretrained and Deterministic}: Maximum softmax probability \cite{hendrycks2016baseline}
     of $N$ networks trained independently on each dataset using either random initialization or pre-trained weights from Sports1M \cite{karpathy2014large}. (We set $N = 5$ for each method below.)
    \item \textbf{Monte-Carlo Dropout} (\textit{MC Dropout}):  Dropout activated at test time as an  approximate bayesian inference in deep Gaussian processes \cite{gal2016dropout}.
    \item \textbf{Temperature Scaling}\footnote{https://github.com/gpleiss/temperature\_scaling} (\textit{TempScaling}):  Post-hoc calibration of softmax probability by temperature scaling using a validation set \cite{guo2017calibration}.
    \item \textbf{Last Layer Dropout} (\textit{LL Dropout}): Bayesian inference for the parameters of the last layer only: Dropout activated at test time on the activations before the last layer.
    \item \textbf{Last Layer Stochastic Variational Bayesian Inference} (\textit{LL SVI}): Mean field stochastic variational inference on the last layer using Flipout \cite{wen2018flipout}.
    \item \textbf{Ensembling} (\textit{Ens}): Average prediction of three networks trained independently on each training set using pre-trained weights \cite{lakshminarayanan2017simple}. Similarly to Table \ref{table:resens}, we use ensembling as a plausible approximation of one model's robustness for real-world scenarii.   
\end{itemize}


\subsection{Discussion}

We present the results obtained by probabilistic methods for both evaluation protocols: train-test and ensembling on Table \ref{tab:final}. This allows us to report the effect of dataset shift on accuracy and calibration for the probabilistic deep learning methods. Naturally, we would like to obtain a model, that is well-calibrated on the training and testing distributions of each dataset and remains calibrated with ensembling. We observe that, similarly to the deterministic methods, the quality of predictions consistently degrades with dataset shift regardless of the selected probabilistic method for both $PIE$ and $JAAD_{all}$. However, overall robustness degrades more significantly for some methods. For instance, TempScaling, \textit{e.g.} post-hoc calibration of softmax probability, seems to be one of the best train-test probabilistic methods in regards to expected calibration error (ECE) when evaluated in a standard train-test procedure but falls behind when evaluated under dataset shift. In fact, when evaluated under dataset shift, all the methods except Non-pretrained ones outperform TempScaling in regards to ECE. Similarly, we report that better calibration and accuracy on each test set does not correlate with better calibration under ensembling: the average ECE of the methods when evaluated with classical train-test scenario is [0.166, 0.074, 0.056, 0.042, 0.079, 0.070] and the average ECE of the same methods under dataset shift are [0.096, 0.077, 0.078, 0.085, 0.077, 0.042]. Interestingly, most of the selected probabilistic methods perform better on average than the deterministic I3D under train-test evaluation protocols but fail to generalize when exposed to dataset shift. The exception to the rule is LL SVI, which looks very promising in terms of generalization to small domain shift. As our experiments required pre-trained weights from I3D, we could not replace each convolutional layer with mean-field variational Flipout layers, we only changed the last layer of the given model to obtain a variational bayesian inference for a quick baseline. Nevertheless, we believe that this could be a future research to consider. We should explore the effects of transferring initially learned features on large bases further away from the target task and explore how probabilistic methods react to transfer-learning and domain-shift.

\section{Conclusion}
In this paper, we show that the classical train-test sets evaluation for pedestrian crossing prediction, \textit{i.e.}, models being trained and tested on the same dataset, is not sufficient to efficiently compare nor conclude anything about their applicability in a real-world scenario: the benchmarks being either too small or too loose in variety of scenarii, it is easy for a given model to over-fit on a specific target dataset. In order to evaluate the generalization capacity of the approaches, we conduct a study based on direct cross-dataset evaluation for eleven methods representing the diversity of architectures and modalities used for pedestrian crossing prediction. We found a huge lack of generalization and robustness for all selected approaches. This led us to a ranking of existing approaches that is much more complex and less absolute than the standard one. We secondly discuss the importance of quantifying a model's uncertainty. Although this is currently completely disregarded, it is common sense to use it in our field of application. We discover two interesting properties: pre-training well-established models on diverse and dense datasets further away from the target domain before fine-tuning to our target task improves calibration and, two models with equivalent classification scores do not necessarily have equivalent calibration scores. This may prove interesting to consider when comparing their usefulness in real-world scenarii with inputs distribution frequently shifted from the training distribution. Finally, we enforce the importance of evaluating the robustness of pedestrian crossing behavior models by evaluating how trustworthy are their uncertainty estimates under domain shifts with cross-dataset evaluation. We encourage the community to consider those new protocols and metrics in order to reach the end-goal of pedestrian crossing behavior predictors: vehicle implementation. 
In order to build the foundation on which future work should be based on, and, in addition to the eleven deterministic baselines evaluated under domain shift, we report the results of multiple baselines from the probabilistic deep learning literature, designed to tackle the problem of improving model calibration. Given all of the above, we advise the community to change the direction in which we are taking the research field: with so little existing data, non-existent generalization of models, and inconclusive ranking of them, we need to agree to properly evaluate our approaches in order to minimize the noise of our productions and thus, make the research field more sustainable and representative of the real advances to come.


%





\ifCLASSOPTIONcaptionsoff
  \newpage
\fi





\bibliographystyle{IEEEtran}
\bibliography{IEEEabrv,Bibliography}

\vfill


\end{document}